%% file: main.tex
\documentclass[10pt,twocolumn,letterpaper]{article}

\usepackage[pagenumbers]{cvpr}      %

\input{preamble}

\input{sections/numbers}

\newif\ifincludesupplementary
\includesupplementaryfalse

\definecolor{cvprblue}{rgb}{0.21,0.49,0.74}
\usepackage[breaklinks,colorlinks,allcolors=cvprblue]{hyperref}

\title{ShapeQuest: Towards Granularity-Controlled Edit-Conditioned 3D Asset Retrieval}

\author{%
Faris Abu-Ali$^{1,2,*}$\qquad
Sankalp Sinha$^{1,2,*,\dagger}$\qquad
Didier Stricker$^{1,2}$\qquad
Muhammad Zeshan Afzal$^{1,2}$\\
$^{1}$German Research Center for Artificial Intelligence (DFKI)\qquad
$^{2}$RPTU Kaiserslautern-Landau\\
{\tt\small \{faris.abu\_ali, sankalp.sinha, didier.stricker, muhammad\_zeshan.afzal\}@dfki.de}\\
{\small $^{*}$Equally contributing first authors. \qquad
$^{\dagger}$Corresponding author and project lead.}\\[-22pt]
}

\makeatletter\makeatother
\begin{document}
\twocolumn[{%
\renewcommand\twocolumn[1][]{#1}%
\maketitle
\begin{center}
  \centering
  \captionsetup{type=figure}
  \includegraphics[width=\linewidth]{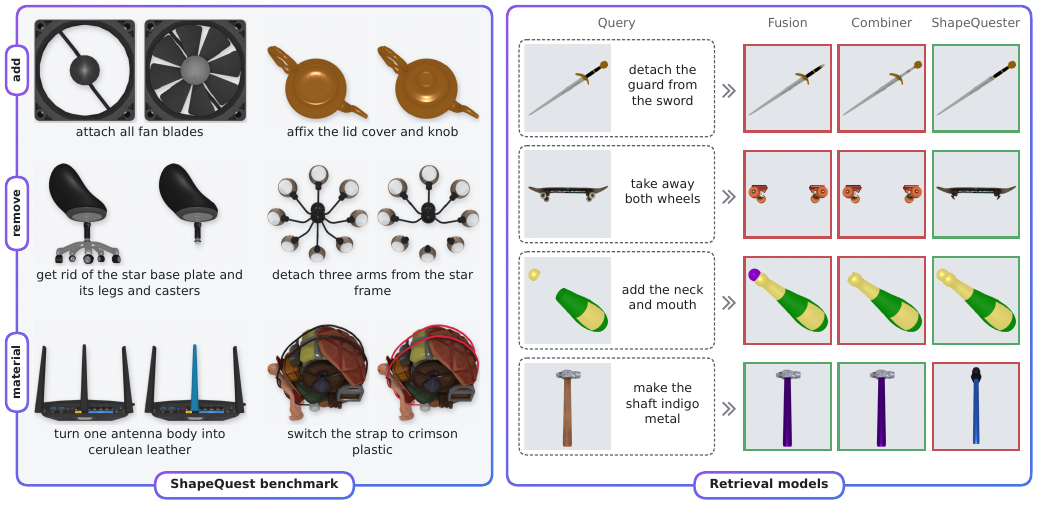}
  \captionof{figure}{\textbf{\ours.}
  \emph{Left:} example triplets, each a source shape, an instruction that adds or removes a part or changes its material, and the target shape.
  \emph{Right:} qualitative top-1 results from the fusion and combiner baselines and \model (green: correct, red: wrong).
  }
  \label{fig:teaser}
\end{center}%
}]

\input{sections/0_abstract}

\input{sections/1_intro}
\input{sections/2_related_work}
\input{sections/3_benchmark}

\input{sections/4_method}
\input{sections/5_experiments}

\input{sections/6_conclusion}

{
    \small
    \bibliographystyle{ieeenat_fullname}
    \bibliography{main}
}

% \ifincludesupplementary
%   \input{sections/supplementary/supplementary}
% \fi

\end{document}

%% file: preamble.tex
\usepackage{xspace}
\xspaceaddexceptions{\%}
\usepackage{multirow}
\usepackage{pifont}
\usepackage{colortbl}   %
\usepackage{makecell}
\usepackage{arydshln}   %
\newcommand{\ours}{ShapeQuest\xspace}
\newcommand{\model}{ShapeQuester\xspace}  %

\newcommand{\ptokens}{patch tokens\xspace}

\newcommand{\headlinemetric}{operation-macro\xspace}
\newcommand{\headlinemetricstacked}[1][]{\makecell{#1operation-\\macro}}
\newcommand{\sbhit}{same-base hit@1\xspace}
\usepackage{fontawesome5}
\definecolor{iconyes}{HTML}{55A868}     %
\definecolor{iconno}{HTML}{606060}      %
\definecolor{iconhead}{HTML}{4C72B0}    %
\definecolor{referencerow}{HTML}{BFE8FA}   %
\definecolor{titlerow}{HTML}{EFEFEF}       %
\definecolor{deltaup}{RGB}{0,110,45}     %
\definecolor{deltadown}{RGB}{175,35,35}   %
\newcommand{\gain}[1]{\,$_{\textcolor{deltaup}{\uparrow#1}}$}
\newcommand{\drop}[1]{\,$_{\textcolor{deltadown}{\downarrow#1}}$}
\newcommand{\gainfull}[1]{\textcolor{deltaup}{$\uparrow$#1}}
\newcommand{\dropfull}[1]{\textcolor{deltadown}{$\downarrow$#1}}

\newcommand{\sbhitstacked}{\makecell{same-base\\hit@1}}   %
\newcommand{\blocktitle}[2]{\multicolumn{#1}{l}{\cellcolor{titlerow}\textit{#2}}}
\newcommand{\titlerule}{\specialrule{\lightrulewidth}{\aboverulesep}{0pt}}
\definecolor{sdgray}{HTML}{4A4A4A}
\definecolor{groupruledots}{HTML}{585858}   %
\newcommand{\sdsize}{\scriptsize}
\newcommand{\sd}[1]{\,{\sdsize\color{sdgray}$\pm$#1}}
\newlength{\grouprulesep}
\newcommand{\grouprule}[1]{\noalign{\vskip\grouprulesep}\arrayrulecolor{groupruledots}\cdashline{#1}[0.6pt/1.6pt]\arrayrulecolor{black}\noalign{\vskip\grouprulesep}}

\newcommand{\cmark}{\textcolor{iconyes}{\faIcon{check-circle}}}
\newcommand{\xmark}{\textcolor{iconno}{\faIcon{times-circle}}}
\newcommand{\pmark}{\textcolor{iconyes}{\faIcon{adjust}}}   %
\newcommand{\headicon}[1]{\textcolor{iconhead}{\faIcon{#1}}~}
\newcommand{\inT}{\textsf{T}}
\newcommand{\inP}{\textsf{P}}
\newcommand{\inI}{\textsf{I}}

\usepackage{listings}
\usepackage{tcolorbox}
\definecolor{placeholder}{HTML}{C44E52}

\tcbuselibrary{listings,breakable}
\newcounter{prompt}

\newtcbinputlisting{\textbox}[3][]{%
  listing file={#3}, listing only, breakable,
  listing options={basicstyle=\fontencoding{T1}\fontfamily{lmtt}\selectfont\footnotesize, upquote=true,
                   aboveskip=0pt, belowskip=0pt, breaklines=true,
                   breakatwhitespace=false, columns=fullflexible, keepspaces=true, frame=none,
                   xleftmargin=0pt, xrightmargin=0pt,
                   moredelim={[s][\color{placeholder}\bfseries]{\{\{}{\}\}}}, #1},
  fonttitle=\footnotesize, title={\hypersetup{allcolors=white}#2}}
\newcommand{\boxcaption}[1]{\if\relax\detokenize{#1}\relax\else{\par\vspace{-2pt}\noindent\small #1\par}\fi\medskip}

\definecolor{codeKeyword}{HTML}{4C72B0}   %
\definecolor{codeComment}{HTML}{2E7D46}   %
\definecolor{codeString}{HTML}{8172B3}    %
\lstdefinestyle{pythoncode}{language=Python, showstringspaces=false,
  keywordstyle=\color{codeKeyword}\bfseries, commentstyle=\color{codeComment}\itshape,
  stringstyle=\color{codeString}, backgroundcolor=\color{gray!4}}
\usepackage{placeins}
\usepackage{environ}   %
\usepackage{etoc}
\AtBeginDocument{\etocdepthtag.toc{body}}   %
\makeatletter\def\sup@dims#1#2#3#4#5{\expandafter\gdef\csname sup@dim@#1\endcsname{{#2}{#3}{#4}{#5}}}\makeatother
\AtEndPreamble{%
  \crefname{appendix}{Sec.}{Secs.}\Crefname{appendix}{Section}{Sections}%
  \crefname{subappendix}{Sec.}{Secs.}\Crefname{subappendix}{Section}{Sections}%
  \crefname{subsubappendix}{Sec.}{Secs.}\Crefname{subsubappendix}{Section}{Sections}%
  \crefname{lstlisting}{Listing}{Listings}\Crefname{lstlisting}{Listing}{Listings}%
  \crefname{prompt}{Prompt}{Prompts}\Crefname{prompt}{Prompt}{Prompts}%
}
\DeclareRobustCommand{\stageref}[1]{\hyperlink{stage#1}{#1}}   %
\newcommand{\stagetarget}[1]{\hypertarget{stage#1}{}}

%% file: sections/numbers.tex
\newcommand{\nCategories}{50\xspace}               %

\newcommand{\nCameras}{17\xspace}
\newcommand{\nValid}{80,298\xspace}                %
\newcommand{\pctValid}{71.9\xspace}

\newcommand{\pctHumanPrecision}{95.5\xspace}       %
\newcommand{\nHumanReviewed}{500\xspace}           %
\newcommand{\nMultiCandidates}{20,466\xspace}   %
\newcommand{\nMultiAfterGates}{12,303\xspace}   %
\newcommand{\nMultiQueries}{12,053\xspace}

\newcommand{\nSingleQueries}{75,771\xspace}        %

\newcommand{\nAddMultiQueries}{3,618\xspace}       %

\newcommand{\nQueries}{87,824\xspace}              %
\newcommand{\nShapes}{126,417\xspace}              %
\newcommand{\nOriginals}{22,630\xspace}            %

\newcommand{\nRemoveQ}{16,871\xspace}
\newcommand{\nAddQ}{12,696\xspace}
\newcommand{\nBalanced}{59,276\xspace}             %
\newcommand{\nMaterialBalanced}{29,709\xspace}     %
\newcommand{\nSingleBalanced}{49,237\xspace}       %
\newcommand{\nMultiBalanced}{10,039\xspace}         %
\newcommand{\nViews}{12\xspace}                    %
\newcommand{\nPoints}{10,000\xspace}               %

\newcommand{\nValBalanced}{15,747\xspace}
\newcommand{\nTestBalanced}{15,684\xspace}         %
\newcommand{\nTrainBalanced}{27,845\xspace}        %

\newcommand{\nGallery}{50,919\xspace}
\newcommand{\nGalleryTargets}{40,896\xspace}
\newcommand{\nGalleryDistractors}{10,023\xspace}
\newcommand{\nGalleryObjects}{8,548\xspace}
\newcommand{\avgDistractors}{5.36\xspace}          %

%% file: sections/0_abstract.tex
\begin{abstract}
Given a reference asset and an instruction such as “remove one armrest,” a retriever must find variants that make the edit and preserve the other parts. This search is well established for images but remains underexplored for individual 3D assets. To study this problem, we introduce ShapeQuest and ShapeQuester, a new benchmark and model for edit-conditioned 3D asset retrieval. ShapeQuest’s open-source pipeline edits textured meshes through part additions, removals, and material and colour changes. It controls edit granularity across five levels, defined by the fraction of the object’s surface area each edit affects. It constructs multiple valid targets when an instruction leaves the choice of part open. A vision-language model judges each edit from views chosen to expose the edited parts, and human raters audit its judgments. ShapeQuest has 87,824 queries in 50 categories and a 50,919-shape gallery with competing edits. To distinguish these variants, ShapeQuester learns retrieval alongside reconstruction of the edited point cloud, with part masks supervising changed and preserved regions. With a frozen pretrained point encoder, ShapeQuester reaches 50.4 operation-macro mAP@10 on the test split, against 47.2 for the fusion baseline with the same inputs and 48.6 for the strongest baseline. Reconstruction adds 6.5 points on the validation split when the point encoder is trained from scratch, including 3.3 from part masks.
\end{abstract}

%% file: sections/1_intro.tex
\section{Introduction}
\label{sec:intro}

A 3D designer furnishing a virtual office finds a chair with the right shape and upholstery, but its armrests keep it from sliding under the desk. A text-based search for ``armless chair'' leaves out the details that made this chair a good choice. With the chair as a reference, the designer can simply ask for ``this one without the armrests.''

For images, this way of searching is known as composed image retrieval (CIR).
A reference image supplies the visual details and a short instruction describes the change~\cite{vo2019tirg,wu2021fashioniq,liu2021cirr}.
In 3D, closely related work includes 3D-CER~\cite{wu_cr-refiner_2026}, which retrieves indoor rooms given an edit to one of their objects.
\emph{Edit-conditioned 3D asset retrieval} remains underexplored.
This task requires finding variants that make the requested part changes while preserving the other parts, even when the gallery contains competing edits of the same asset. Yet no benchmark evaluates this task.

To address this gap, we introduce \textbf{\ours{}}, to the best of our knowledge the first dedicated benchmark for edit-conditioned retrieval of individual 3D assets (\cref{fig:teaser}). We build it by editing named parts of textured meshes from PartNeXt~\cite{wang_partnext_2026}.

\ours{} controls \emph{edit granularity} across five levels, defined by the fraction of the reference object's surface affected by an edit. On the same chair, we can vary the scale of the requested change from a single caster to the entire backrest while preserving the other parts. \ours{} also provides \emph{multiple positive targets} when an instruction leaves the choice of part open. On the same chair, ``remove one armrest'' can yield two valid variants and ``change one leg to dark wood'' can yield four. We construct each valid variant, so mAP rewards models for ranking the full set of valid targets near the top of the gallery.

We build baselines for \ours{} by adapting late fusion and the CLIP4Cir combiner~\cite{baldrati2022clip4cir} from CIR.
For the skateboard query ``take away both wheels,'' both baselines rank wheel assemblies above the board without its wheels (\cref{fig:teaser}).
Retrieval labels identify valid targets but give no direct guidance on which parts should change or remain.
Motivated by these failures, we introduce \textbf{\model{}}, which conditions PointBERT's patch tokens on the instruction and learns retrieval alongside reconstruction of the edited point cloud. \ours{}'s part masks supervise the changed and preserved regions, and we discard the decoder at inference.
\model{} outperforms both the fusion and combiner baselines.
When the point encoder is trained from scratch, part masks account for roughly half the retrieval gain from reconstruction, showing \ours{}'s value.

Our main contributions are summarised as follows:
\begin{itemize}
    \item We introduce \textbf{\ours{}}, a new benchmark for edit-conditioned 3D asset retrieval with five granularity levels and multiple valid targets for underspecified edits.
    \item We develop an open-source pipeline that turns part-annotated meshes into queries with visually audited edits and masks for changed and preserved regions.
    \item We propose \textbf{\model{}}, which uses part masks to reconstruct edited shapes during training. Experiments show that it outperforms the baselines.
\end{itemize}

%% file: sections/2_related_work.tex
\section{Related Work}
\label{sec:related}

\begin{table*}[t]
  \centering
  \footnotesize
  \setlength{\tabcolsep}{3.5pt}
  \newcommand{\twoline}[2]{\begin{tabular}[c]{@{}c@{}}#1\\#2\end{tabular}}
  \newcommand{\twolinel}[2]{\begin{tabular}[c]{@{}l@{}}#1\\#2\end{tabular}}
  \begin{tabular}{llllcccc}
    \toprule
     & & \multicolumn{3}{c}{\headicon{search}Query} & \multicolumn{3}{c}{\headicon{clipboard-check}Evaluation} \\
    \cmidrule(lr){3-5}\cmidrule(lr){6-8}
    work & output & reference & query text & \twoline{change}{applies to} & \twoline{ranked}{gallery} & multi-GT & \twoline{target =}{reference + change} \\
    \titlerule
    \blocktitle{8}{2D retrieval} \\
    FashionIQ~\cite{wu2021fashioniq}, CIRR~\cite{liu2021cirr} & 2D images & image & change & whole image & \cmark & \xmark & \xmark \\
    CIRCO~\cite{baldrati_zero-shot_2023} & 2D images & image & change & whole image & \cmark & \cmark & \xmark \\
    COR~\cite{wang2025cor} & \twolinel{object mask}{in a given image} & \twolinel{image +}{object mask} & change & object & \xmark & \cmark & \xmark \\
    \titlerule
    \blocktitle{8}{3D editing: generates a new shape} \\
    ShapeTalk~\cite{achlioptas_shapetalk_2023} & edited shape & shape & shape difference & whole shape & \xmark & \xmark & \xmark \\
    ShapeWalk~\cite{slim_shapewalk_2024} & edited shape & shape & change & program param & \xmark & \xmark & \cmark\rlap{$^{*}$} \\
    3DEditVerse~\cite{xia_towards_2025} & edited shape & shape & change & image region & \xmark & \xmark & \xmark \\
    Pxform~\cite{weng_feedforward_2026} & edited shape & shape & change & part & \xmark & \xmark & \pmark\rlap{$^{\dagger}$} \\
    \titlerule
    \blocktitle{8}{3D retrieval} \\
    Text2Shape~\cite{chen2018text2shape} & 3D shapes & -- & caption & -- & \cmark & \xmark & -- \\
    3DCoMPaT200~\cite{ahmed_3dcompat200_2025} & 3D shapes & -- & \twolinel{part/material}{description} & -- & \cmark & \xmark & -- \\
    ROOMELSA~\cite{nguyen2025roomelsa} & 3D shapes & \twolinel{masked}{room image} & description & -- & \cmark & \xmark & -- \\
    3D-CER~\cite{wu_cr-refiner_2026} & 3D rooms & room & change & object in room & \cmark & \cmark & \xmark \\
    \textbf{\ours (ours)} & \textbf{3D shapes} & \textbf{shape} & \textbf{change} & \textbf{part} & \cmark & \cmark & \cmark \\
    \bottomrule
  \end{tabular}
    \caption{\textbf{\ours beside the closest work.}
  Only \ours combines a \emph{ranked gallery}, several correct answers per query (\emph{multi-GT}), and an answer that is the reference with exactly the requested change (\emph{target = reference + change}).
  The \emph{query text} either describes the target or asks for a change to the reference.
  $^{*}$Procedurally generated shapes only. $^{\dagger}$Removals and additions only.}
  \label{tab:related}
\end{table*}

\noindent\textbf{Composed image retrieval.}
Composed image retrieval (CIR) ranks a gallery of images given a reference image and a text that asks for a change~\cite{vo2019tirg}.
CLIP4Cir~\cite{baldrati2022clip4cir,baldrati_composed_2023} showed that a combiner over frozen CLIP~\cite{radford2021clip} features is a strong baseline, and many methods refine it~\cite{saito2023pic2word,gu_compodiff_2023,wang_generative_2025,li_imagine_2025,liu_candidate_2024,chen_offset_2025,udhayanan_learning_2023,huynh_collm_2025}.
FashionIQ~\cite{wu2021fashioniq} and CIRR~\cite{liu2021cirr} pair reference images with hand-written changes, CIRCO~\cite{baldrati_zero-shot_2023} adds several correct answers per query, and recent benchmarks control the requested change~\cite{byun_macir_2025,song_edir_2026} or test whether a model needs both query inputs~\cite{attimonelli_circus_2026}.
COR~\cite{wang2025cor} and i-CIR~\cite{psomas_instance_2025} narrow the change from a whole image to an object or an instance.
All of this work concerns images, where a change applies to the whole picture or to an object in it.
\ours brings the task to 3D shapes, where the change names one of the shape's parts, and \cref{tab:related} compares it with the closest work.
We port the CIR late-fusion and combiner designs from 2D to 3D as baselines.

\noindent\textbf{Language-guided 3D editing.}
ShapeTalk~\cite{achlioptas_shapetalk_2023} and LADIS~\cite{huang_ladis_2022} study shape changes described in language, ShapeWalk~\cite{slim_shapewalk_2024} chains synthetic edits of procedural shapes, and recent datasets broaden paired supervision for feed-forward editors~\cite{xia_towards_2025,cai_native_2025,xu_beyond_2026,liyi_omni-3dedit_2026,li_alchemy3d_2026}.
Pxform~\cite{weng_feedforward_2026} grounds its edits in semantic parts, the image-guided EditVerse3D~\cite{yin_editverse3d_2026} supervises edited and preserved regions, and EditHero~\cite{yu_edithero_2026} builds sequences of part edits.
3DEditVerse~\cite{xia_towards_2025} and Pxform regenerate most targets with a 3D generator, so regions outside the edit can change too.
All of these works generate one edited shape and score it against a single reference.
\ours evaluates part edits by retrieval: a model ranks a gallery of existing shapes, and each target is built on the original mesh, so only the named part changes.
When the change names one of several symmetric parts, such as one leg of a chair, changing any of them gives a correct answer.

\noindent\textbf{3D shape retrieval.}
Text2Shape~\cite{chen2018text2shape} embeds text and shapes jointly, and ULIP~\cite{xue2023ulip,xue2024ulip2}, OpenShape~\cite{liu_openshape_2023}, and Uni3D~\cite{zhou2024uni3d} align point-cloud encoders with CLIP at scale, while Cap3D~\cite{luo2023cap3d} and MARVEL-40M+~\cite{sinha2025marvel} caption 3D shapes automatically.
ROOMELSA~\cite{nguyen2025roomelsa} retrieves the shape of an object masked in a room image, given a description of it.
MetaFind~\cite{pan_metafind_2025} retrieves shapes from any mix of text, image, and 3D queries, guided by the layout of the scene they are placed into.
3DCoMPaT200~\cite{ahmed_3dcompat200_2025} retrieves shapes from a text that names their parts and their materials.
3D-CER~\cite{wu_cr-refiner_2026} retrieves rooms from a reference room and a change to one of its objects, with several correct rooms mined by programmatic predicates.
Each of these queries either describes the target shape or asks for a change to one object of a room.
\ours retrieves single shapes, and its answer is the reference itself with exactly the requested part change, so a retriever must tell it apart from the source and from the other edits of the same object.

\noindent\textbf{Part-annotated 3D data.}
ShapeNet~\cite{chang2015shapenet} provides categorised meshes, and PartNet~\cite{mo2019partnet} annotates their hierarchical parts.
The 3DCoMPaT family~\cite{li20223dcompat,slim_3dcompat_2025,ahmed_3dcompat200_2025} assigns materials to parts for recognition and retrieval, and PartNeXt~\cite{wang_partnext_2026} combines fine-grained hierarchical parts with textured meshes.
These datasets annotate the parts a shape has, while a composed retrieval benchmark needs changes to those parts.
\ours builds these changes from PartNeXt: its pipeline sizes every part, applies part edits to the original meshes, and validates each edit, so PartNeXt's part annotations become queries, answers, and the masks that supervise our \model model.

%% file: sections/3_benchmark.tex
\section{Our Benchmark: \ours}
\label{sec:benchmark}

We first define the retrieval task, the benchmark protocol, and the evaluation metrics (\cref{sec:task}).
We then present the dataset construction and validation pipeline (\cref{sec:construction}).
Further details are provided in \cref{sec:supplementary_benchmark}.

\subsection{Task Definition and Metrics}
\label{sec:task}
\noindent\textbf{Retrieval task definition.}
Let $\mathcal{G}$ be a gallery of 3D shapes.
A query $(s,q)$ pairs a source shape $s$ with a natural-language edit instruction $q$.
Each instruction is \emph{part-grounded}: it names a part or group of parts.
It specifies one of three operations: addition (add), removal (remove), or a change of material and colour (material).
Edit granularity is the fraction of the original object's surface area affected by the requested change.
The correct answer set $\mathcal{A}(s,q)\subseteq\mathcal{G}$ contains all shapes obtained by programmatically applying the edit specified by $q$ to $s$, while leaving all other parts unchanged.
The retrieval task is to rank $\mathcal{G}$ so that these targets appear first.
The \emph{single-ground-truth} track has $|\mathcal{A}|=1$, while the \emph{multi-ground-truth} track has $|\mathcal{A}|\geq 2$.
Multiple targets are valid when the instruction leaves the choice among interchangeable parts unspecified.
For example, ``remove one leg'' on a chair with four interchangeable legs has four correct targets, each with one leg removed.
\looseness=-1

\noindent\textbf{Benchmark protocol and release.}
We divide queries into training, validation, and test sets, grouping them by PartNeXt object UUID so that all variants of an object remain in one split.
Validation and test share a gallery containing all source and target shapes from both splits.
The query's own source remains a candidate, and source shapes that answer no query serve as same-object distractors for every query of their object (\cref{fig:stage7_gallery_composition,fig:stage7_distractors_per_query}).
Together with the other edited versions of the same object, these candidates require the retriever to distinguish the requested change.
Each shape is provided as \nViews views rendered from a Fibonacci sphere and a coloured point cloud of \nPoints points.
Points are sampled in proportion to surface area, and each cloud is normalised to the unit sphere (\crefrange{fig:stage7_released_views}{fig:stage7_released_point_clouds}).
Training and evaluation use the operation-balanced query subsets constructed in Stage~\stageref{7}; selecting these subsets leaves the gallery unchanged.
We publish validation answers for local evaluation and keep test answers private for scoring through an evaluation server.
\noindent\textbf{Metrics.}
Our headline score is \textbf{\headlinemetric mAP@10}, reported in \cref{tab:main}.
It averages mAP@10 over add, remove, and material, giving each operation equal weight.
mAP@10 rewards ranking correct targets early within the top ten results and accounts for all valid targets of each query~\cite{baldrati_zero-shot_2023}.
We also report mAP@10 separately for geometry (add and remove) and material edits.
Finally, \textbf{same-base hit@1} measures how often a correct target ranks first among gallery shapes derived from the source's underlying object.
This assesses edit discrimination with object identity fixed.
\Cref{sec:supplementary_protocol_and_metrics} gives every formula.
\begin{figure*}[t]
  \centering
  \includegraphics[width=\linewidth]{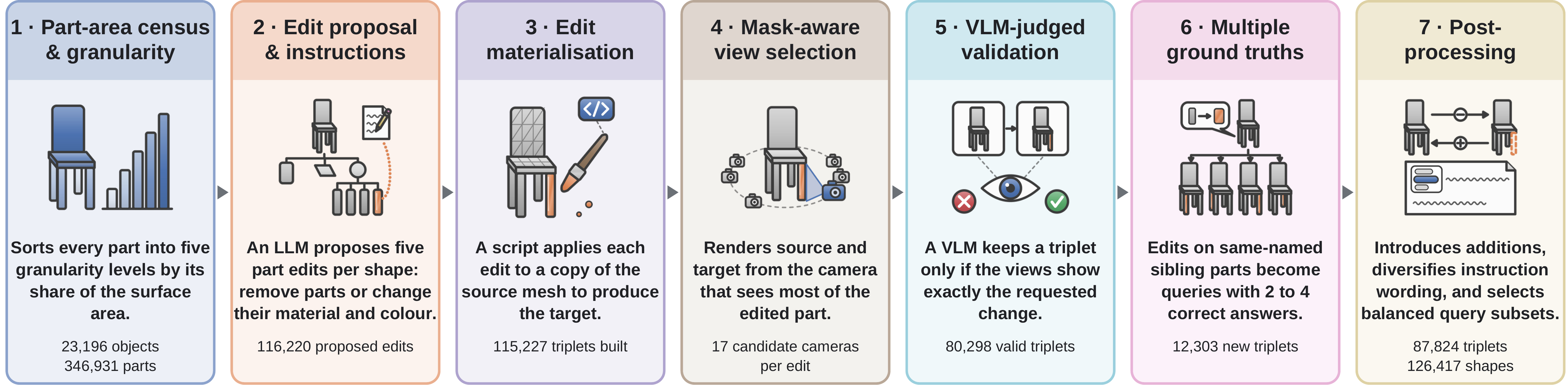}
  \caption{\textbf{The \ours construction pipeline.}
  The seven progressive stages turn PartNeXt shapes into an edit-conditioned 3D asset retrieval benchmark. Queries span five edit granularity levels and can have multiple correct targets. \Cref{sec:supplementary_construction_pipeline} details every stage.
  }
   
  \label{fig:pipeline}
\end{figure*}

\subsection{Construction Pipeline}
\label{sec:construction}
Our first contribution is the pipeline in \cref{fig:pipeline}, which builds \ours from PartNeXt.
Stage \stageref{1} measures every part and assigns it a granularity level; Stages \stageref{2} and \stageref{3} propose part edits with an LLM and apply them to the meshes; Stages \stageref{4} and \stageref{5} render every edit and validate it with a VLM; Stage \stageref{6} adds queries with several correct answers; and Stage \stageref{7} produces the released splits and gallery.

\noindent\stagetarget{1}\textbf{1 · Part-area census and granularity calibration.}
We measure each annotated part's surface area as a fraction of its shape's total surface area.
We use this distribution to define five granularity levels with approximately equal numbers of parts.
In order of increasing relative size, these are \emph{extra fine}, \emph{fine}, \emph{medium}, \emph{coarse}, and \emph{extra coarse} (\cref{fig:stage1_granularity_ladder}).
Mesh preprocessing removes non-object geometry that would distort the area measurements (\cref{fig:stage1_studio_floor_plane}).
Calibration and preprocessing details are provided in \cref{sec:supplementary_stage1_part_area_census}.

\noindent\stagetarget{2}\textbf{2 · Edit proposal and instruction authoring.}
An LLM reads each shape's part hierarchy and the granularity labels assigned in Stage~\stageref{1}.
It is asked to propose five edits spanning the available granularity levels, each with an instruction naming the affected parts.
At this stage, edits either remove parts or change their material and colour.
The material choices are \emph{wood}, \emph{metal}, \emph{plastic}, \emph{leather}, and \emph{glass}, each with predefined metallicity and roughness ranges.
Each shape also receives a colour shortlist drawn from a predefined palette to diversify the proposed colours (\cref{fig:stage2_colour_diversity}).
Since an edit can affect several parts, we measure its extent by their combined share of the shape's surface area.
We recalibrate the five granularity levels on these edit areas so that each label reflects the full affected region (\cref{fig:stage1_granularity_queries_one_shape}).
We discard removals that would delete the whole shape (\cref{fig:stage2_empty_scene_removals}).
Instructions that identify a part only as \texttt{Other} are rewritten when the intended part is clear and discarded otherwise (\cref{fig:stage2_other_part_violations}).

\noindent\stagetarget{3}\textbf{3 · Edit materialisation.}
We construct each target programmatically using a deterministic Python script that edits a copy of the source shape's glTF file.
For removals, the script deletes the faces of the selected parts.
For material edits, it assigns those parts the proposed material and colour.
We convert proposed colours from sRGB to linear RGB before writing the material values (\cref{fig:stage3_srgb_to_linear}).
We also retain existing vertex colours (\cref{fig:stage3_vertex_colours}).
Edits that cannot be applied are discarded.
Each target therefore differs from its source only in the selected parts and keeps the geometry, material, and colour of all other parts.

\noindent\stagetarget{4}\textbf{4 · Mask-aware view selection.}
We select a camera view for each edit to make the change visible during validation.
We evaluate \nCameras candidate poses: four azimuths at each of four elevations, plus one pose along the edited part's mean surface normal (\cref{fig:stage4_candidate_cameras}).
Using PartNeXt's annotations, we render a mask of the edited region on the source mesh and count its visible pixels at each pose.
We select the pose that shows the most of the edited region while keeping the shape recognisable.
We then move the camera closer to make the edit easier to inspect (\cref{fig:stage4_fixed_view_vs_mask_aware_view}).
Edits with insufficient visibility from all candidate poses are withheld (\cref{fig:stage4_unshowable_edits}).
Further selection details and worked examples are provided in \cref{sec:supplementary_stage4_mask_aware_view_selection}.
Finally, we render the source and target from the same selected camera (\cref{fig:stage4_one_shape_every_edit}), providing a matched image pair for validation in Stage~\stageref{5}.

\begin{table*}[t]
  \centering
  \footnotesize
  \begin{minipage}[b]{0.55\linewidth}
    \centering
    \includegraphics[height=5.3cm]{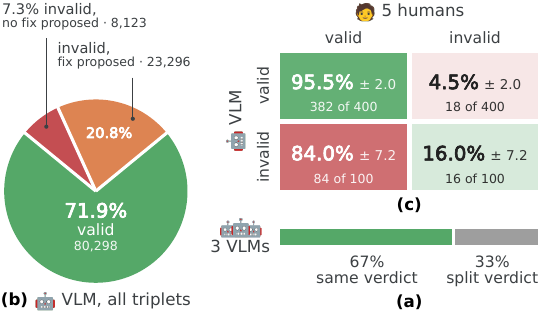}
  \end{minipage}\hfill
  \begin{minipage}[b]{0.42\linewidth}
    \raggedleft
    \setlength{\tabcolsep}{2.5pt}
    \begin{tabular}[b]{@{}lr@{}}
      \toprule
      \multicolumn{2}{@{}l}{\headicon{cubes}\textbf{\ours at a glance}} \\
      \midrule
      triplets                            & \nQueries \\
      \quad single-GT / multi-GT          & \nSingleQueries\ / \nMultiQueries \\
      triplets (operation-balanced)       & \nBalanced \\
      \quad single-GT / multi-GT          & \nSingleBalanced\ / \nMultiBalanced \\
      \quad material / remove / add & \nMaterialBalanced\ / \nRemoveQ\ / \nAddQ \\
      \quad train / val / test            & \nTrainBalanced\ / \nValBalanced\ / \nTestBalanced \\
      shapes (released meshes)            & \nShapes \\
      \quad PartNeXt objects they come from & \nOriginals \\
      categories                          & \nCategories \\
      per shape                           & \nViews views + \nPoints-pt cloud \\
      gallery shapes (val + test)         & \nGallery \\
      \quad PartNeXt objects they come from & \nGalleryObjects \\
      \quad targets / same-object distractors & \nGalleryTargets\ / \nGalleryDistractors \\
      \quad distractors per triplet (mean) & \avgDistractors \\
      \bottomrule
    \end{tabular}
  \end{minipage}
  \par\smallskip
  \begin{minipage}[t]{0.55\linewidth}
    
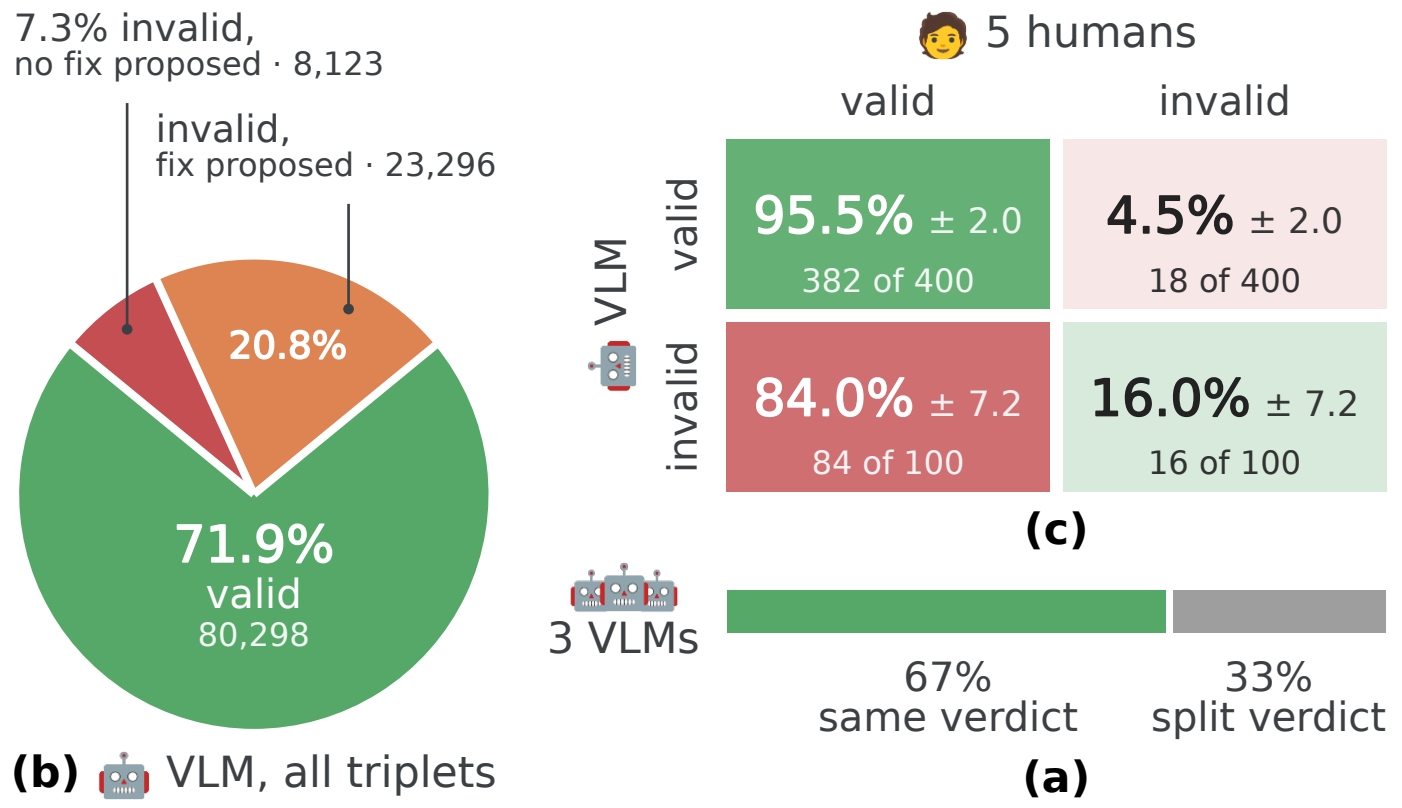
\captionof{figure}{\textbf{Triplet quality.}
    (a)~Three VLM judges concur on 67\% of the triplets of 100 shapes.
    (b)~The chosen VLM judge accepts \pctValid\% of all triplets.
    (c)~Five humans each rated the same \nHumanReviewed triplets blind to the VLM's verdicts; by majority vote, they concur on 95.5\% of those it accepted. 
    }
    \label{fig:human_validation_matrix}
  \end{minipage}\hfill
  \begin{minipage}[t]{0.42\linewidth}
    \captionof{table}{\textbf{\ours in numbers.}
    We train and evaluate all models on the operation-balanced triplets; \cref{tab:supplementary_construction_funnel} describes the construction funnel for \ours{}.}
    \label{tab:stats}
  \end{minipage}
\end{table*}

\noindent\stagetarget{5}\textbf{5 · Triplet validation and human evaluation.}
We use InternVL3-78B as the VLM judge, following a comparison of three VLMs (\cref{fig:stage5_judge_selection}).
It accepts a triplet only if the visible change in the paired views from Stage~\stageref{4} matches the instruction.
It explains each verdict and may suggest a revised instruction for rejected triplets (\cref{fig:stage5_judge_examples,fig:stage5_more_rejected_edits}). These suggestions are kept separate from the benchmark. Five human reviewers independently assessed the same \nHumanReviewed{} triplets without seeing the VLM's verdicts.
The sample included accepted and rejected triplets at all five granularity levels.
By majority vote, the human reviewers confirmed \pctHumanPrecision\% of sampled acceptances overall and more than 90\% at every level, including extra-fine edits (\cref{fig:human_validation_matrix}; \cref{fig:human_validation_sheet}). The human reviewers also agreed closely with one another (\cref{fig:human_validation_sheet_part2}).
When the human majority disagreed with the VLM, the humans usually judged a rejected triplet valid.
Together, these findings support the retained triplets' quality across edit sizes and indicate cautious filtering by the VLM judge.

\noindent\stagetarget{6}\textbf{6 · Multiple ground truths.}
We mine the \nValid triplets accepted in Stage~\stageref{5} for edits of two or more parts with the same PartNeXt name, which yields \nMultiCandidates candidates.
Three checks keep \nMultiAfterGates of them: (1) the edit covers every such part, (2) these parts have similar surface areas, and (3) they number two to four.
A rule-based template rewrites each instruction to request the edit on one of them, and the script of Stage~\stageref{3} builds one target per part: ``change all legs to gold metal'' on a four-legged table becomes ``change one leg to gold metal'', with four targets.
\Crefrange{fig:stage6_diagram}{fig:stage6_multigt_categories} and \crefrange{lst:stage6_step1_mine}{lst:stage6_step3_expand} give examples and code.

\noindent\stagetarget{7}\textbf{7 · Post-processing.}
We introduce additions by reversing half the single-ground-truth removals, swapping source and target and rewriting the instruction to request the missing part (\cref{fig:stage7_diagram}). Each mesh pair appears in only one direction.
If an addition leaves the part's material unspecified, all gallery targets restoring the part and preserving the rest are correct; \nAddMultiQueries additions have several correct answers.
We diversify opening verbs with a fixed pool of hand-checked phrasings, including ``detach'' and ``take off'' for removals (\cref{tab:stage7_verb_pools}).
Finally, material edits outnumber removals (\cref{fig:stage1_query_distribution}), so we subsample Stage~\stageref{6}'s material edits to match its removals. Each split's balanced subset contains all geometry queries and a similar number of material ones.

%% file: sections/4_method.tex
\section{Our Model: \model}
\label{sec:method}

\begin{figure*}[t]
  \centering
  \includegraphics[width=\linewidth]{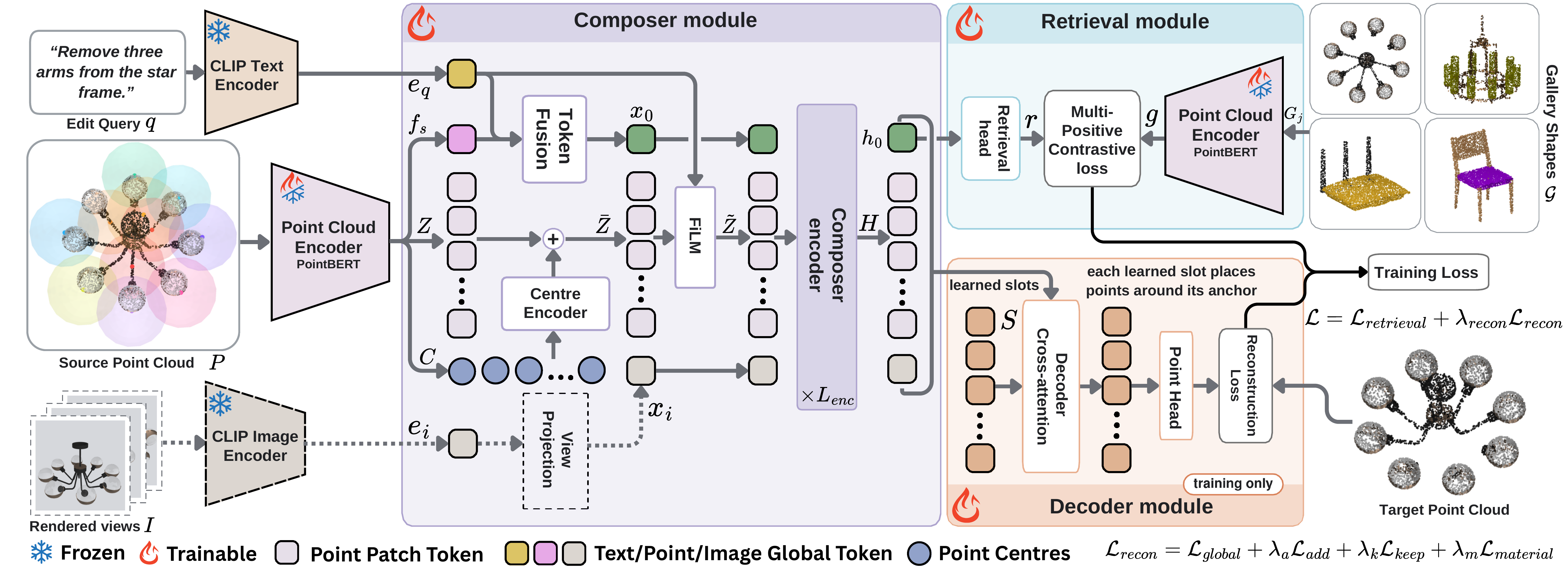}
  \caption{\textbf{\model.}
  The instruction modulates the source patch tokens, which a transformer mixes with a CLS token to form the retrieval embedding $r$.
  Gallery shapes are encoded by the same PointBERT and ranked by cosine similarity.
  A training-only point decoder predicts the edited shape under \ours's part masks; the ground-truth target illustrates its output.
  PointBERT is frozen with pretrained weights and trained jointly in the from-scratch setting.}
  \label{fig:model}
\end{figure*}

\noindent\textbf{Overview and Motivation.}
Retrieval labels tell the model which targets match a query, but give no direct guidance on which parts should change or remain.
For ``remove three arms from the star frame'' (\cref{fig:model}), a valid target has three fewer arms and retains the rest of the frame.
We use auxiliary reconstruction to encourage the composer to encode this edit and the preserved regions in the pooled retrieval vector.
\model builds on PointBERT and CLIP text and image encoders.
The Composer Module conditions source patch tokens on the instruction and combines them with global shape features and, optionally, view features.
The Retrieval Module maps the result to a query embedding and ranks gallery shapes.
During training, the Decoder Module reconstructs the edited cloud from the composed tokens.
We supervise this reconstruction with \ours's part masks on changed and preserved regions. We discard the decoder at inference.

\noindent\textbf{Encoders.}
We encode the source point cloud $P$ with PointBERT~\cite{yu2022pointbert}.
We either use OpenShape's pretrained weights~\cite{liu_openshape_2023} and keep the encoder frozen, or train it from scratch (\cref{sec:exp_setup}).
We retain its projected global feature $f_s \in \mathbb{R}^{D}$ and the patch tokens from its final transformer layer, $Z \in \mathbb{R}^{M \times d}$.
Keeping both outputs gives the composer local shape features alongside global object context.
Each patch token $z_j$ has a centre $c_j \in \mathbb{R}^{3}$, and we stack these centres as $C \in \mathbb{R}^{M \times 3}$.
The frozen OpenCLIP ViT-bigG-14 text tower~\cite{cherti2023openclip} encodes the instruction $q$ as $e_q \in \mathbb{R}^{D}$.
The corresponding frozen image tower optionally encodes one of the rendered source views $I$ as $e_i \in \mathbb{R}^{D}$.
We $\ell_2$-normalise the global point and CLIP features.
\Cref{sec:supplementary_implementation} gives the encoder settings and layer sizes.

\noindent\textbf{Composer Module.}
We condition patch tokens $Z$ on the instruction to help the composer represent local changes before forming a single retrieval embedding.
Centre encoding and feature-wise linear modulation (FiLM)~\cite{perez2018film} produce $\bar Z$ and $\tilde Z$, respectively, with rows
\begin{equation}
\begin{aligned}
  \bar z_j
  &= \mathrm{LayerNorm}(z_j) + \mathrm{MLP}(c_j), \\
  \tilde z_j
  &= \bar z_j \odot (1+\tanh\gamma) + \beta.
\end{aligned}
\label{eq:shapequester_film}
\end{equation}
A single linear map of $e_q$ produces $\gamma,\beta \in \mathbb{R}^{d}$, shared across patches.
Centre encoding tells the composer where patches lie, while FiLM scales and shifts their channels according to the instruction.

We prepend a CLS token, $x_0=W_s f_s+W_q e_q$, which carries the global shape context and the instruction.
An optional view enters as another token, $x_i=W_i e_i+v_i$, where $v_i$ is a learned view-type embedding.
A stack of $L_{\mathrm{enc}}$ transformer encoder layers lets these tokens exchange information through self-attention.
After layer normalisation, it produces $H \in \mathbb{R}^{L \times d}$, with $L=M+1$, or $M+2$ when a view is included.
The output CLS token $h_0$ combines the global context with the instruction-conditioned patch features.
Additional view configurations are described in \cref{sec:supplementary_rendered_view}.

\noindent\textbf{Retrieval Module.}
The retrieval head maps $h_0$ to the query embedding
$r=\mathrm{norm}(W_r h_0) \in \mathbb{R}^{D}$,
where $\mathrm{norm}$ denotes $\ell_2$ normalisation.
The same PointBERT encodes each gallery shape from its point cloud alone.
We normalise its global feature $f$ as
$g=\mathrm{norm}(f) \in \mathbb{R}^{D}$
and rank shapes by cosine similarity $r^\top g$.

Training uses query--target pairs, with one pair for each correct target.
Let $B$ be the batch size and $\mathcal P_i$ the indices of all batch candidates that are valid targets for query $i$.
With scores $s_{ij}=r_i^\top g_j/\tau$ and fixed temperature $\tau$, we use the multi-positive contrastive loss~\cite{oord2018cpc,khosla2020supcon}:
\begin{equation}
\begin{gathered}
  \ell_i =
  -\frac{1}{|\mathcal P_i|}
  \sum_{j\in\mathcal P_i}
  \log
  \frac{\exp(s_{ij})}
       {\sum_{k=1}^{B}\exp(s_{ik})},
  \\[4pt]
  \mathcal L_{\mathrm{retrieval}} =
  \frac{\sum_i K_i^{-1}\ell_i}
       {\sum_i K_i^{-1}}.
\end{gathered}
\label{eq:shapequester_retrieval}
\end{equation}
The multi-positive contrastive loss in \cref{eq:shapequester_retrieval} averages contrastive terms over each query's valid targets in the batch, regardless of which example supplied them.
The weight $K_i^{-1}$ corrects for the query's repetition across training pairs, where $K_i$ counts all its correct targets in \ours.

\noindent\textbf{Decoder Module.}
The decoder reconstructs the edited cloud to guide the shared composer on both the requested change and the parts to preserve.
Learned slots $S$ read $H$ through cross-attention, and a point head predicts a coarse cloud of $N$ points, $\hat P \in \mathbb{R}^{N \times 6}$, with XYZ coordinates and RGB colour.
For queries with several correct targets, we select one fixed canonical target $P_t$ for reconstruction, giving the deterministic decoder a consistent answer.
All correct targets remain positives for retrieval.
\Cref{sec:supplementary_point_decoder} details the decoder architecture.

We compare prediction and target in the source cloud's normalised coordinate frame.
The global reconstruction loss in \cref{eq:shapequester_global_reconstruction} combines symmetric Chamfer distance~\cite{fan2017pointset} in XYZ with colour error:
\begin{equation}
  \mathcal L_{\mathrm{global}}
  =
  \mathrm{CD}_{\mathrm{xyz}}(\hat P,P_t)
  +
  \lambda_c\,\mathrm{CD}_{\mathrm{rgb}}(\hat P,P_t).
  \label{eq:shapequester_global_reconstruction}
\end{equation}
Both terms average errors in the two directions.
The XYZ errors are squared distances; the RGB errors are squared colour differences at the corresponding Euclidean XYZ nearest neighbours.

The part masks add supervision on the regions named by the instruction and those that should remain unchanged.
$\mathcal L_{\mathrm{add}}$ covers added target parts using absolute position and squared RGB errors.
$\mathcal L_{\mathrm{keep}}$ covers unchanged parts in both source and target using squared position and RGB errors.
$\mathcal L_{\mathrm{material}}$ supervises the target RGB on recoloured parts.
Each masked region has its own average, giving small edited parts a separate contribution beyond the whole-cloud loss.
Removal needs no additional term because global reconstruction penalises predicted geometry absent from the target.

We average the reconstruction components over canonical examples and train both tasks with the joint loss in \cref{eq:shapequester_joint_loss}:
\begin{equation}
\begin{gathered}
  \begin{aligned}
    \mathcal L_{\mathrm{recon}}
    &=
    \mathcal L_{\mathrm{global}}
    + \lambda_a\mathcal L_{\mathrm{add}}
    \\
    &\quad+
    \lambda_k\mathcal L_{\mathrm{keep}}
    + \lambda_m\mathcal L_{\mathrm{material}},
  \end{aligned}
  \\[4pt]
  \mathcal L =
  \mathcal L_{\mathrm{retrieval}}
  + \lambda_{\mathrm{recon}}\mathcal L_{\mathrm{recon}}.
\end{gathered}
\label{eq:shapequester_joint_loss}
\end{equation}
The add and material terms apply only to their respective operations.
\Cref{sec:supplementary_point_decoder} gives full reconstruction details, and \cref{sec:supplementary_loss_weights} reports loss weights and their ablations.
Masks enter only the training loss.
At inference, we discard the decoder and retrieve shapes from the source cloud, instruction, and optional rendered view.

%% file: sections/5_experiments.tex
\section{Experiments}
\label{sec:experiments}

\begin{table*}[t]
  \centering
  \small
  \renewcommand{\sdsize}{\fontsize{6}{7}\selectfont}   %
  \setlength{\tabcolsep}{3.4pt}   %
  \setlength{\grouprulesep}{0.75pt}   %
  \begin{tabular}{@{}clcrccccc|c@{}}
    \toprule
    \makecell{PointBERT\\pretrained} & model & query & trainable & \makecell{point\\grouping} & \headlinemetricstacked & geometry & material & \makecell{same-base\\hit@1} & \headlinemetricstacked[test ] \\
    \midrule
    \cmark & sum (zero-shot) & \inT\,\inP & 0 & original & 33.55 & 25.16 & 50.34 & 13.88 & 33.56 \\
    \midrule
    \multirow{10}{*}{\xmark} & \multirow{2}{*}{fusion} & \multirow{2}{*}{\inT\,\inP} & \multirow{2}{*}{45.4\,M} & original
        & 43.27\sd{0.91} & 32.91\sd{1.03} & 64.73\sd{1.06} & 48.57\sd{0.40} & 42.89\sd{0.75} \\
      & & & & nearest & 45.22\sd{0.43} & 34.38\sd{0.42} & 67.59\sd{0.75} & 49.95\sd{0.73} & 44.87\sd{0.53} \\
    \grouprule{5-10}
      & \multirow{2}{*}{fusion} & \multirow{2}{*}{\inI\,\inT\,\inP} & \multirow{2}{*}{50.4\,M} & original
        & 43.29\sd{0.73} & 33.00\sd{0.50} & 64.54\sd{1.12} & 48.81\sd{0.39} & 42.98\sd{0.61} \\
      & & & & nearest & 44.71\sd{0.56} & 34.11\sd{0.41} & 66.55\sd{0.93} & 50.46\sd{0.66} & 44.49\sd{0.62} \\
    \grouprule{5-10}
      & \multirow{2}{*}{combiner-\inI} & \multirow{2}{*}{\inI\,\inT\,\inP} & \multirow{2}{*}{111.0\,M} & original
        & 43.97\sd{0.60} & 33.23\sd{0.51} & 66.18\sd{0.91} & 49.00\sd{0.42} & 43.87\sd{0.55} \\
      & & & & nearest & 45.81\sd{0.13} & 34.91\sd{0.17} & 68.21\sd{0.18} & 50.60\sd{0.68} & 45.78\sd{0.43} \\
    \grouprule{5-10}
      & \multirow{2}{*}{\model} & \multirow{2}{*}{\inT\,\inP} & \multirow{2}{*}{42.5\,M} & original
        & 45.15\sd{0.17} & 35.27\sd{0.32} & 65.89\sd{1.01} & 47.61\sd{0.37} & 44.67\sd{0.18} \\
      & & & & nearest & 46.89\sd{0.48} & 36.15\sd{0.43} & 69.40\sd{1.10} & 50.05\sd{0.84} & 46.53\sd{0.83} \\
    \grouprule{5-10}
      & \multirow{2}{*}{\model} & \multirow{2}{*}{\inI\,\inT\,\inP} & \multirow{2}{*}{43.2\,M} & original
        & 46.73\sd{0.40} & 35.27\sd{0.40} & 70.67\sd{0.92} & 51.83\sd{0.30} & 46.28\sd{0.59} \\
      & & & & nearest & \textbf{48.77}\sd{0.25} & \textbf{37.33}\sd{0.42} & \textbf{72.70}\sd{1.03} & \textbf{53.75}\sd{0.96} & \textbf{48.49}\sd{0.19} \\
    \midrule
    \multirow{5}{*}{\cmark} & fusion              & \inT\,\inP       & 13.1\,M & \multirow{5}{*}{original} & 46.75\sd{0.09} & 33.82\sd{0.12} & 73.58\sd{0.34} & 50.47\sd{0.61} & 47.23\sd{0.22} \\
      & fusion              & \inI\,\inT\,\inP & 18.0\,M & & 47.08\sd{0.28} & 34.55\sd{0.45} & 73.12\sd{0.06} & 51.91\sd{0.26} & 47.26\sd{0.20} \\
      & combiner-\inT       & \inI\,\inT\,\inP & 78.7\,M & & 48.23\sd{0.11} & 36.64\sd{0.08} & 72.54\sd{0.47} & 49.31\sd{0.37} & 48.64\sd{0.29} \\
      & \model             & \inT\,\inP       & 10.2\,M & & \textbf{50.72}\sd{0.27} & \textbf{38.27}\sd{0.31} & \textbf{76.71}\sd{0.33} & 55.77\sd{0.64} & \textbf{50.44}\sd{0.13} \\
      & \model              & \inI\,\inT\,\inP & 10.9\,M & & 49.78\sd{0.30} & 37.39\sd{0.19} & 75.82\sd{0.74} & \textbf{56.35}\sd{0.70} & 49.77\sd{0.29} \\
    \bottomrule
  \end{tabular}
  \caption{\textbf{Main comparison on \ours} (mAP@10 and \sbhit in \%, mean $\pm$ std over three seeds; val split, last column test split).
  Bold marks the best in a block. The sum row adds the source's OpenShape embedding to the instruction's CLIP text embedding and trains nothing.
  Inputs (query): instruction \inT, point cloud \inP, rendered view \inI; \crefrange{tab:full_metrics_test_operation_macro}{tab:full_metrics_val_overall} list every model on every metric.}
  \label{tab:main}
\end{table*}

\subsection{Experimental Setup}
\label{sec:exp_setup}

\noindent\textbf{Point encoder and point grouping.}
We evaluate PointBERT in two settings: frozen with OpenShape's pretrained weights~\cite{liu_openshape_2023}, and trained from scratch with the rest of the retriever.
The CLIP towers remain frozen in both settings.
Since \ours's edits can affect small parts, we also compare two ways of forming local patches.
Both rules select up to 64 points within the same radius of each patch centre.
During training, we shuffle the point clouds with the same policy and seeds for both rules.
The \emph{original} rule takes the first 64 points in the shuffled input order, while the \emph{nearest} rule selects the 64 closest to the centre, forming a more compact neighbourhood.
The frozen pretrained encoder retains its original grouping, while the from-scratch models are trained with both rules.
We compare retrievers only within the same encoder setting and grouping.
\Cref{sec:supplementary_point_grouping} describes additional details of point grouping.

\noindent\textbf{Baselines.}
We adapt two established designs from composed image retrieval to \ours: \emph{late fusion} and the \emph{CLIP4Cir combiner}.
The \emph{fusion} baseline concatenates the global features of its inputs and maps them to a query through an MLP, following the concatenation baseline evaluated in the TIRG study~\cite{vo2019tirg}.
This lets the network learn interactions between the input features.
The \emph{combiner} follows CLIP4Cir~\cite{baldrati2022clip4cir}, which has shown strong performance on both FashionIQ and CIRR.
To accommodate our three input modalities, we first fuse two inputs and then compose the result with the third.
A learned gate balances these two features, while a residual MLP adds their learned composition.
Both baseline families use the encoder settings, point-only gallery encoder, and retrieval objective described in \cref{sec:method}.
Their query networks operate on global features and are trained with retrieval supervision alone.
\Cref{tab:main} lists the inputs used by each model.

\noindent\textbf{Training.}
We train all models for 10 epochs on the operation-balanced training split, using a single NVIDIA L40S GPU.
We use AdamW~\cite{loshchilov2019adamw} with a base learning rate of $10^{-4}$ and a batch size of 32.
All models use the retrieval loss of \cref{sec:method}, with an additional reconstruction loss for \model.
For each run, we keep the checkpoint with the highest validation \headlinemetric mAP@10.
For the main comparisons, we report the mean and sample standard deviation over three runs with seeds 0, 42, and 84.
Additional training details are given in \cref{sec:supplementary_training}.

\subsection{Main Results}
\label{sec:exp_main}

\noindent\textbf{Do CIR-inspired designs carry over to 3D shapes?}
Yes, OpenShape already places shapes and text in one embedding space, which gives \ours a training-free reference: sum-\inT\inP, the first row of \cref{tab:main}, adds the instruction's text embedding to the source's shape embedding.
Trained on \ours with the same frozen encoders, fusion and the combiner score \gainfull{13.20} to \gainfull{14.68} over this reference, and the combiner leads fusion in every block.
Neither reaches 50 \headlinemetric mAP@10.

\noindent\textbf{Does \model outperform the baselines?}
It scores the highest \headlinemetric mAP@10 in every block of \cref{tab:main}, in both regimes and with both groupings.
Against fusion-\inT\inP, which reads the same two inputs, it scores \gainfull{3.97} with the pretrained PointBERT and \gainfull{1.67} from scratch with the nearest grouping; \cref{sec:supplementary_paired_tests} lists the paired differences behind every lead.
Reading only the instruction and the point cloud, it also leads the combiner, with a fraction of the trainable parameters and without a rendered view.
On \sbhit the pretrained \model leads as well; from scratch, \model with a view leads while the one without does not.
The leads hold on the held-out test split as well.
\Crefrange{fig:qualitative_overview}{fig:qualitative_model_comparison} show qualitative results on the val split.

\subsection{Ablation Studies}
\label{sec:exp_ablation}

\begin{table}[t]
  \centering
  \footnotesize   %
  \renewcommand{\sdsize}{\scriptsize}   %
  \setlength{\tabcolsep}{1pt}   %
  \begin{tabular}{lllll}
    \toprule
    composer & reconstruction & $\lambda_{\text{recon}}$ & \headlinemetricstacked & \sbhitstacked \\
    \midrule
    global feature & none & 0 & 45.87\sd{0.52}\drop{1.02} & 49.41\sd{0.53}\drop{0.64} \\
    global feature & $\mathcal{L}_{\text{recon}}$ & 1 & \textbf{46.76}\sd{0.43}\drop{0.13} & \textbf{53.41}\sd{0.64}\gain{3.36} \\
    \midrule
    \ptokens & none & 0 & 40.42\sd{0.51}\drop{6.47} & 33.19\sd{0.76}\drop{16.86} \\
    \ptokens & $\mathcal{L}_{\text{global}}$ & 1 & 43.56\sd{0.79}\drop{3.33} & 40.47\sd{2.21}\drop{9.58} \\
    \rowcolor{referencerow}\ptokens & $\mathcal{L}_{\text{recon}}$ & 1 & \textbf{46.89}\sd{0.48} & \textbf{50.05}\sd{0.84} \\
    \midrule
    \ptokens & $\mathcal{L}_{\text{recon}}$ & 0.1 & 41.89\sd{1.50}\drop{5.00} & 37.01\sd{3.65}\drop{13.04} \\
    \bottomrule
  \end{tabular}
  \caption{\textbf{Ablation of \model} (PointBERT from scratch, nearest grouping): the \ptokens need the reconstruction loss.
  The blue row is the full model and the reference of the arrows; bold marks the best of a column. Pretrained (\cref{tab:models_composer_and_reconstruction}), the \ptokens lead the global feature by \gainfull{1.77}.}
  \label{tab:ablation}
\end{table}

\noindent\textbf{Effect of \ptokens and the reconstruction loss.}
\Cref{tab:ablation} ablates the two design choices of \model with PointBERT trained from scratch: whether the composer reads PointBERT's \ptokens or its global feature alone, and which reconstruction loss it trains with.
We first train without the reconstruction loss.
The \ptokens then score \dropfull{5.45} against the global feature, and we attribute this to the missing training signal: the composer can attend to every patch, but nothing tells it which patches the instruction concerns.
We then add the loss back in two steps.
The full $\mathcal{L}_{\text{recon}}$ scores \gainfull{6.47} over no reconstruction loss: $\mathcal{L}_{\text{global}}$ alone gives \gainfull{3.14}, and the three mask terms, $\mathcal{L}_{\text{add}}$, $\mathcal{L}_{\text{keep}}$, and $\mathcal{L}_{\text{material}}$, another \gainfull{3.33}.
Without the \ptokens, that is, with the global feature alone, the same loss gives only \gainfull{0.89}; with the full $\mathcal{L}_{\text{recon}}$, the global feature therefore scores \dropfull{0.13} against the \ptokens on \headlinemetric mAP@10, while it stays \gainfull{3.36} ahead on \sbhit.
The weight of the loss matters as well: the last row keeps every term of $\mathcal{L}_{\text{recon}}$ but divides it by ten, and the gain over no reconstruction loss shrinks from \gainfull{6.47} to \gainfull{1.47}.
With the pretrained PointBERT the loss matters less (\cref{tab:models_composer_and_reconstruction}). \Cref{sec:supplementary_further_analysis,sec:supplementary_mask_terms,sec:supplementary_loss_weights} analyse the encoder settings, a rendered view, LoRA, the composer, the mask terms, and the loss weights.
\Cref{tab:models_every_metric} breaks every model down by operation, edit size, and granularity level.

%% file: sections/6_conclusion.tex
\label{sec:conclusion}

\section{Limitations and Conclusion}
In \ours{}, edits are limited to removing, adding, and recolouring labelled parts, and a recoloured part receives one of five constant materials and no texture map (\cref{sec:supplementary_materials_and_textures}). OpenShape's pretraining data overlaps PartNeXt's sources (\cref{sec:supplementary_further_analysis}).
Geometry edits, additions and removals, score far below material edits for every model, and \cref{sec:supplementary_same_base} analyses how often \model ranks the unedited source before a correct target.

\ours{} makes edit-conditioned 3D asset retrieval measurable through part-grounded instructions, five granularity levels, and several valid targets.
Our open-source pipeline constructs targets by editing source meshes while preserving the other parts, and the gallery keeps both sources and competing edits.
\model{} conditions PointBERT's \ptokens on the instruction and uses the pipeline's part masks in reconstruction training, outperforming the fusion and combiner baselines of \cref{tab:main} with fewer trainable parameters.